\documentclass[10pt, conference, letterpaper]{IEEEtran}

\usepackage{cite}
\usepackage{amsmath,amssymb,amsfonts}
\usepackage{graphicx}
\usepackage[font=footnotesize]{subfig}
\usepackage{textcomp}
\usepackage{xcolor}
\usepackage{algorithm}
\usepackage[noend]{algpseudocode}
\usepackage{subfloat}
\usepackage[colorinlistoftodos]{todonotes}
\def\BibTeX{{\rm B\kern-.05em{\sc i\kern-.025em b}\kern-.08em
    T\kern-.1667em\lower.7ex\hbox{E}\kern-.125emX}}

\allowdisplaybreaks   

\newcommand{\ignore}[1]{}

\begin{document}

\title{Online Learning for Adaptive Probing and Scheduling in Dense WLANs}

\author{
    \IEEEauthorblockN{Tianyi Xu$^{a}$, Ding Zhang$^{b}$, Zizhan Zheng$^{a}$}
    \IEEEauthorblockA{$^a$ Department of Computer Science, Tulane University, New Orleans, LA, USA}
    \IEEEauthorblockA{$^b$ Department of Computer Science, George Mason University, Fairfax, VA, USA \vspace{-15pt}}
}

\newtheorem{observation}{Observation}
\newtheorem{theorem}{Theorem}
\newtheorem{claim}{Claim}
\newtheorem{remark}{Remark}
\newtheorem{corollary}{Corollary}
\newtheorem{lemma}{Lemma}
\newtheorem{assumption}{Assumption}
\newcommand{\red}[1]{{\color{red} #1}}
\newcommand{\blue}[1]{{\color{blue} #1}}
\newcommand\numberthis{\addtocounter{equation}{1}\tag{\theequation}}
\renewcommand{\algorithmicrequire}{\textbf{Input:}}  
\renewcommand{\algorithmicensure}{\textbf{Output:}}



\maketitle

\begin{abstract}

Existing solutions to network scheduling typically assume that the instantaneous link rates are completely known before a scheduling decision is made or consider a bandit setting where the accurate link quality is discovered only after it has been used for data transmission. In practice, the decision maker can obtain (relatively accurate) channel information, e.g., through beamforming in mmWave networks, right before data transmission. However, frequent beamforming incurs a formidable overhead in densely deployed mmWave WLANs. In this paper, we consider the important problem of throughput optimization with joint link probing and scheduling. 
The problem is challenging even when the link rate distributions are pre-known (the offline setting) due to the necessity of balancing the information gains from probing and the cost of reducing the data transmission opportunity. We develop an approximation algorithm with guaranteed performance when the probing decision is non-adaptive and a dynamic programming-based solution for the more challenging adaptive setting. We further extend our solutions to the online setting with unknown link rate distributions and develop a contextual-bandit based algorithm and derive its regret bound. Numerical results using data traces collected from real-world mmWave deployments demonstrate the efficiency of our solutions.  
 
\end{abstract}


\section{Introduction}
An important obstacle to efficient resource allocation in wireless networks is the uncertain and time-varying nature of link quality. Traditionally, it is common to assume that the decision maker can observe the time-varying channel condition at the beginning of a time slot before a scheduling decision is made~\cite{neely2010}, leading to the so called {\it opportunistic scheduling} problems that can be solved using various stochastic optimization techniques. More recently, online learning algorithms have been intensively studied for wireless resource allocation, including link scheduling~\cite{modiano2019UCBgreedy,bo2019sleeping}, rate adaptation~\cite{atilla2019bandit} and beam selection~\cite{slivkins2011contextual} where a typical assumption is that the channel condition of a link is unknown before a scheduling decision is made and is discovered only after the link has been used for data transmission (i.e., the {\it bandit} setting).  

In practice, a network controller can often obtain (relatively) accurate channel conditions before transmitting data. For example, access points (APs) in 60 GHz millimeter-wave (mmWave) WLANs can use beamforming to identify the link quality to its clients.    
However, frequent beamforming between a large number of APs and clients incurs a formidable overhead. For example, it takes approximately 5ms to train the downlink of an AP with 64 Tx sectors to its clients with 16 Rx sectors as shown in~\cite{Ding-hotmobile}, and the overhead increases linearly with the number of APs.
Thus, it is crucial to develop joint beamforming and scheduling schemes that can achieve the optimal tradeoff between the reduction of uncertainty and the extra overhead introduced by beamforming. This is especially important in densely deployed mmWave WLANs that are increasingly being used to provide high capacity and reliability to clients. However, neither traditional stochastic optimization approaches nor bandit-based online learning methods can be directly applied as the former ignores the overhead associated with beamforming while the latter does not utilize the side information provided by beamforming. 

In this work, we propose a joint probing and scheduling framework for throughput optimization. We consider a scenario where a mobile client can probe a subset of APs from all the neighboring APs before picking one of them for data transmission and study both the non-adaptive setting where the set of APs is chosen independently of the probing results as well as the more powerful yet challenging adaptive setting where the next probing decision can depend on the previous probing results. We first consider the offline setting where the distributions of link rates are pre-known. Deriving an optimal probing and scheduling strategy is challenging, even in this case, due to the need to balance the information gained from probing and the cost of reduced data transmission opportunity. 
We design an approximation algorithm for the non-adaptive setting and derive its performance bound. For the adaptive setting, we propose a dynamic programming-based solution and prove that the algorithm is optimal for Bernoulli link rates. 

In the more challenging online setting where the link rate distributions are unknown a priori, we present a novel extension of the contextual bandit learning framework by incorporating adaptive probing into the classic exploration vs. exploitation tradeoff. In particular, the probing results help with both the exploitation in the current round and the exploration that can benefit future rounds. Through a careful analysis of the reward function and by utilizing the results for the offline setting, we show that the UCB-based stochastic combinatorial bandit algorithm in~\cite{chen2016combinatorial} can be adapted to our problem and derive its regret bound for both non-adaptive probing under general distributions and adaptive-probing with Bernoulli link rates.

Our framework can potentially be applied to a broad class of sequential decision making problems where probing can be used to reduce uncertainty before a decision is made. For example, consider the problem of shortest path routing in a road network with unknown traffic. A path searcher can make a limited number of queries to a travel server to obtain hints of real time travel latency~\cite{bhaskara2020adaptive}, before picking a path.
Further, the path searcher may utilize contextual data such as time and weather to assist decision making. This problem can be formulated as a combinatorial bandit problem with probing. 

We have made the following contributions in this paper. 

\begin{itemize}
    \item We propose a joint probing and scheduling framework for throughput optimization in wireless networks.
    \item In the offline setting with known link rate distributions, we develop an approximation solution when the probing decision is non-adaptive and a dynamic programming-based solution for the adaptive setting. We show that both algorithms are optimal for Bernoulli link rates and bound the performance of the former for general distributions. 
    \item In the online setting with unknown link rate distributions, we adapt a stochastic combinatorial bandit algorithm to our problem and derive its regret. 
    \item Our solutions are validated using channel and mobility traces collected from a real testbed. 
\end{itemize}

\section{Related Work}

\noindent{\bf Multi-armed bandits}: The multi-armed bandit (MAB) problem provides a principled framework for sequential decision making under uncertainty~\cite{auer2002finite,zhu2020distributed,landgren2021distributed}. The framework has been extensively applied to various domains, including wireless resource allocation and link scheduling, with multiple variants of the vanilla MAB model considered~\cite{auer2002finite,bubeck2012regret,gittins2011multi}. In~\cite{stahlbuhk2019learning}, the Upper Confidence Bound (UCB) algorithm is integrated with the classic greedy maximal matching algorithm for wireless scheduling. 
In \cite{fml2018}, the problem of beam selection in mmWave based vehicular networks is formulated as a contextual bandit problem with the vehicle’s
direction of arrival and other traffic information as context. In \cite{gupta2019link}, a constrained Thompson Sampling (TS) algorithm is used to solve the link rate selection problem by exploiting the property that a higher data rate is typically associated with a lower transmission success probability. However, none of the above studies consider joint probing and play. Although the recent work~\cite{xu2021joint} considers joint probing and play, it focuses on non-adaptive probing strategies and derives the regret bound for Bernoulli link rates only. 


\vspace{1ex}
\noindent{\bf Decision-making with probes}: Searching for best alternatives with uncertainty using probing is a problem initiated in economics~\cite{pandora-Weitzman} and has found applications in database query optimization~\cite{munagala2005pipelined,deshpande2016approximation,liu2008near} and wireless communications \cite{guha2006optimizing}{\cite{shu2009throughput}}. Various objectives have been considered including searching for extreme values with a limited budget, maximizing the largest
value found minus the total probing cost spent, or solving a knapsack problem with uncertain weights and costs~\cite{goel2006asking}. Since the problems are typically NP-hard problems under general probability distributions, approximation algorithms have been considered for the non-adaptive probing~\cite{goel2006asking} and adaptive probing settings~\cite{guha2007model}. Recently, probing policies have been studied for optimizing path search over a road network~\cite{bhaskara2020adaptive}. However, all the above works consider the offline setting with known distributions. Further, none of them consider the same objective function used in our problem. In~\cite{he2013endhost}, an endhost-based shortest path routing problem with unknown and time-varying link qualities is modeled as an online learning problem. The paper considers decoupled routing and probing decisions, where a single probe is allowed before a route is selected at each time round, and link qualities can only be observed through probes (but not the actual routing decisions). Because of the simplifications, the paper adopts a simple probing strategy by always probing the path containing the least measured link. In contrast, we consider multiple (and possibly adaptive) probes in each round, and both probing and play provide feedback on rewards. Our work is also closely related to the sequential channel sensing and access problem in cognitive radio networks studied in~\cite{li2013almost}. However, the work only considers Bernoulli channels without context. Further, a channel can be accessed only when it is sensed idle. In contrast, our problem allows playing an unprobed arm.


\section{System Model and Problem Formulation}\label{sec:model}

In this section, we present our joint probing and scheduling framework. We start with the general online setting, which includes the offline setting as a special case. We further introduce the concepts of non-adaptive probing and adaptive probing.  


\subsection{Contextual bandits with probing}
\ignore{
\red{should we use the reward evaluated ?
\begin{align*}
R_{e}(Y, S)=
\left\{
             \begin{array}{ll}
              \mu(\mathop{\rm argmax}\limits_{j \in A \backslash S}{\rm UCB}_j), &\mathop{\rm max}Y_{a_i}<\mathop{\rm max}\limits_{j \in A \backslash S}{\rm UCB}_j\\
              \\
              \mathop{\rm max}\limits_{a_i \in S}Y_{a_i},  &\mathop{\rm max}Y_{a_i} \ge \mathop{\rm max}\limits_{j \in A \backslash S}{\rm UCB}
             \end{array}
\right.
\end{align*}}
}

Consider a set of APs that are connected to a high-speed backhaul to collectively serve a set of mobile users. AP collaboration helps boost wireless performance in both indoor and outdoor environments and is especially useful for directional mmWave networks that are susceptible to blockage~\cite{fml2018,Ding-hotmobile}. For simplicity, we assume that the beamforming process determines the best Tx and Rx beam sectors and do not distinguish AP selection from beam selection. Our framework readily applies to the more general setting of joint AP and beam selection. 

To simplify the problem, we focus on the single client setting in this work. 
Let $X$ be a set of contexts that correspond to the location (or a rough estimate of it) of a moving client. 
Let $A$ be a discrete set of arms that correspond to the set of APs, and $N \triangleq |A|$. We consider a fixed time horizon $T$ that is known to the decision maker. In each time round $t$, the decision maker first receives a context $x_t\in{X}$ and then plays an arm $a_t\in{A}$ and receives a reward $Y_{a_t | x_t} \in [0, 1]$, which is $i.i.d.$ sampled from an {\it unknown} distribution $D_{a_t | x_t}$, that depends on both the context $x_t$ and the arm $a_t$. We assume that the outcomes from all $N$ arms are mutually independent.
In this work, we 
assume that for all context-arm pairs, $D_{a|x}$ is a discrete distribution with a finite support $\{y_{a|x,1},...,y_{a|x,l}\}$ where $0=y_{a|x,1}< \cdots <y_{a|x,l-1}< y_{a|x,l} = 1$. Let $\mathcal{Y}$ denote the union of possible values across all the arms. We have $|\mathcal{Y}| \leq N \times l$. 
Let $\mu_{a|x}$ denote the expected value of $D_{a|x}$. The finite support assumption is mainly needed for our offline dynamic programming algorithm (Algorithm~\ref{alg_new}) and the online algorithm (Algorithm~\ref{alg:tswg2}). Let $\mathbb{D} = \{D_{a|x}\}$ denote the collection of distributions over all the context-arm pairs. The context sequence $(x_t)_{t \in \mathbb{N}}$ is considered to be outside the decision making process. In the AP selection problem, the reward corresponds to the data rate that a client at a certain location can receive from an AP. 


In the classical stochastic bandit setting, the instantaneous reward of an arm is not revealed until it is played. 
In contrast, we consider a more general setting where after receiving the context $x_t$, the network controller can first probe a subset of arms $S_t \subseteq A$, observe their outcomes $Y_t = \{Y_{a|x_t}: a \in S_t\}$, where $Y_{a|x_t}$ is an $i.i.d.$ sample of $D_{a|x_t}$, then pick an arm $a_t$ to play (which may be different from the set of probed arms). If $a_t$ is from the probing set, the reward received is the same as the probed value. That is, we assume that probing is accurate and the channel condition stays the same within a time round. To model the probing overhead, we further assume that up to $K$ arms can be probed in a single time round and the (normalized) throughput obtained for probing $S_t$ with $|S_t| = k$ and playing $a_t$ is 
\begin{align}
R(S_t,Y_t,a_t|x_t) = \frac{K-k}{K} Y_{a_t|x_t}.
\end{align}

\noindent Without loss of generality, we assume $K \leq N$. The goal of the network controller is to maximize the total expected throughput over the time horizon $T$. In general, the optimal choice of $a_t$ given $S_t$ depends on $Y_t$. Let $r(S_t,a_t|x_t) \triangleq \mathbb{E}_{Y_t}[R(S_t,Y_t,a_t|x_t)]$.


\subsection{Non-adaptive probing}
We distinguish two types of probing strategies in this work. In the non-adaptive case, the probing set $S_t$ is chosen independently of the probing results. 
This setting is useful when adaptive probing is infeasible but it also serves as an important baseline. In this case, the decision in each time round becomes a pair $(S_t,a_t)$ where $S_t$ is the set of arms to probe and $a_t$ is the arm to play. Let $H_{t-1}$ denote the history observations that the decision maker has received before time $t$. Then the probing policy $\pi_1$ maps $H_{t-1}$ to a subset of arms $S_t$ to probe while the play policy $\pi_2$ maps $(H_{t-1},S_t, Y_t)$ to the arm $a_t$ to play. Let $\pi=(\pi_1,\pi_2)$. 

Let $(S^*_t,a^*_t)$ denote the optimal offline non-adaptive solution under context $x_t$ when the true distributions $D_{a|x}$ are known for all context-arm pairs. The goal of maximizing the expected cumulative reward then converts to minimizing the expected regret $\mathbb{E}(\textsf{Reg}(T))$ with $\textsf{Reg}(T)$ defined as 
\begin{align}
\textsf{Reg}(T)=\sum_{t=1}^T (r(S_t^\star,a^{\star}_t| x_t) - r^\pi(S_t,a_t| x_t))\label{def:regret}
\end{align}

\noindent where $r^\pi$ denotes the throughput obtained by following policy $\pi$ and the expectation is with respect to the randomness in both outcomes and policies.
The goal is to obtain a policy $\pi$ with sublinear regret where $\lim_{T \rightarrow \infty}\mathbb{E}(\textsf{Reg}(T))/T = 0$.

For general context-arm distributions $D_{a|x}$, it can be computationally difficult to find the optimal solution even when the link rate distributions are perfectly known (the offline setting). On the other hand, efficient approximation algorithms may exist. In particular, we say that an offline algorithm is an $\alpha$-approximation $(0< \alpha \le 1)$ if it obtains an expected throughput that is at least $\alpha \mathbb{E}r(S^{\star},a^{\star}|x)$ for any context $x \in X$. In this case, it is difficult to obtain sublinear regret with respect to~\eqref{def:regret}. Instead, our goal is to obtain sublinear  $\alpha$-approximation regret defined as follows~\cite{chen2016combinatorial}:
\begin{align}
\textsf{Reg}_{\alpha}(T)=\sum_{t=1}^T (\alpha r(S_t^\star,a_t^{\star}| x_t) - r(S_t,a_t| x_t))
\end{align}

\subsection{Adaptive probing}
In the adaptive setting, $S_t$ is chosen sequentially where the next arm to probe can depend on the previous probing results in the current time round. Let $S_{i,t}$ denote the subset of arms probed in the first $i$ probing steps in time $t$, and $Y_{i,t}$ the corresponding probing outcome. Then the probing policy $\pi_1$ can generally be defined as a mapping from $(H_{t-1}, S_{i,t}, Y_{i,t})$ to the next probing decision (either the next arm to probe or terminating probing). The play policy $\pi_2$ has the same form as the non-adaptive case. We can then define regret and $\alpha$-approximation regret similarly to the non-adaptive case by using the optimal offline adaptive solution as the baseline.

\subsection{Assumption on the context space}
We consider contextual bandits with Lipschtz-continuity to handle large context spaces. 
To this end, we normalize the context space to the $[0,1]$ interval so that $X = [0, 1]$ and the expected rewards are assumed to be Lipshitz with respect to the contexts:
\begin{align}
|r(S,a|x)-r(S,a|x')| \leq L|x-x'|, \forall S \subseteq A, a \in A, x, x' \in X
\label{Lipschitzcontext}
\end{align}
where $L$ is the Lipschitz constant known to the algorithm. 

\section{The Offline Case}\label{sec:algo}

In this section, we consider the offline case where the reward distributions $D_{a|x}$ are fully known for all context-arm pairs. In this case, the history $H_{t-1}$ is irrelevant to decision making and we can focus on a single time round. To simplify the notation, we omit the time index $t$ and context $x_t$, and write $S = S_t$, $Y = Y_t$, $Y_a = Y_{a|x_t}$, $D_a = D_{a|x}$, and $\mu_a = \mu_{a|x}$. We first observe that given a probing set $S$ and its outcome $Y$, the optimal arm to play can be easily identified, which allows us to focus on $\pi_1$ in this section.
\begin{observation}
Given a probing set $S$, let $Z_{a} = Y_{a}$ for $a \in S$ and $Z_{a} = \mu_{a}$ for $a \in A \backslash S$. It is optimal to play an arm $a$ that maximizes $Z_{a}$.\label{obs:pi2}
\end{observation}



\subsection{The offline problem with non-adaptive probing}

We first consider the offline problem with non-adaptive probing. Given Observation~\ref{obs:pi2}, the problem boils down to finding a subset of arms $S \subseteq A$ to probe to maximize $R(S) \triangleq \frac{K-|S|}{K}g(S)$ where $g(S)$ is defined as 
\begin{align}
 g(S)=\mathbb {E}_{Y_a\sim D_a}\Big[\max(\max_{a \in S}Y_a, \max_{j \in A\backslash S}\mu_j)\Big]
\end{align}


We first note that a similar problem has been considered in~\cite{goel2006asking}, where given a budget $k$, the goal is to find a subset $S \subseteq A$ with $|S| = k$ such that $g(S)$ is maximized. This problem is NP-hard, yet a simple greedy algorithm gives a constant approximation factor (see Algorithm~\ref{alg:tswg4}). The key idea is to consider a new objective function 
\begin{align}
 f(S)=\mathbb {E}_{Y_a\sim D_a}\Big[\max_{a \in S}Y_a\Big]
\end{align}

\noindent This is equivalent to always choosing a probed arm with maximum reward to play without considering the unprobed arms. $f(S)$ has the following nice properties:

\begin{itemize}
    \item Monotonicity: $f(S) \leq f(T)$ for any $S \subseteq T \subseteq A$; \item Submodularity: $f(S \cup \{a\})-f(S) \geq f(T \cup \{a\}) - f(T)$ for $S \subseteq T \subseteq A$ and $a \not\in T$.
\end{itemize}

The two properties together ensure that a simple greedy algorithm obtains an approximation factor of $\alpha = 1-1/e$ to the problem $\max_{S \subseteq A, |S|=k} f(S)$~\cite{nemhauser1978analysis} . The greedy algorithm starts with $S = \emptyset$ and in each step, a new arm that maximizes the marginal improvement $f(S \cup \{a\})-f(S)$ is added to $S$ until $|S|=k$ (see lines 1-4 in Algorithm~\ref{alg:tswg4}). To obtain a solution to the problem $\max_{S \subseteq A, |S|=k} g(S)$, we simply take the greedy solution $S$ found and compare $f(S)$ with $\max_{a \in A } \mu_a$ and take the maximum (lines 5-7 in Algorithm~\ref{alg:tswg4}). 

\begin{algorithm}[!t]
  \caption{Greedy probing with a fixed number of probes} 
  \label{alg:tswg4}
  \begin{algorithmic}[1]
    \Require
    $A$: a set of $N$ arms; $\mathbb{D}$: reward distributions of all the arms; $k$: a fixed budget for probing
    \Ensure
    $S^{\rm pr}$: the probing set
    \State $S \leftarrow \emptyset$
    \For {$i=1$ to $k$}
       \State $a \leftarrow \mathop{\rm argmax}\limits_{a \in A \backslash S}{(f(S \bigcup\{a\})-f(S))}$
       \State $S \leftarrow S \cup \{a\}$
       \EndFor
     \State $S^{pr} \leftarrow S$
    \State $a \leftarrow \mathop{\rm argmax}\limits_{a \in A}\mu_{a}$
    \If{$f(S) < \mu_a$}
        $S^{pr} \leftarrow \emptyset$
    \EndIf
  \end{algorithmic}
\end{algorithm}

We then present Algorithm~\ref{alg:tswg1} that adapts Algorithm~\ref{alg:tswg4} to our problem of maximizing $R(S)$. The key difference is that Algorithm~\ref{alg:tswg1} maintains the greedy solution $S_k$ for each $k \in \{0,..., K-1\}$ and identifies the best solution among them (with respect to $\frac{K-|S|}{K}f(S)$), which is then compared with the maximum $\mu_a$ among all the arms. We establish the following approximation ratio of our algorithm. 

\begin{theorem}
Let $S^\star$ denote the optimal probing set with respect to the reward distributions $\mathbb{D}$.  Algorithm~\ref{alg:tswg1} outputs a subset $S^{\rm pr}$ such that $R(S^{\rm pr}) \ge \alpha R(S^\star)$ where $\alpha=\frac{e-1}{2e-1}$.
\label{offalpha1}
\end{theorem}
\begin{IEEEproof}
Let $S_k$ be the $k$-th probing set found by Algorithm~\ref{alg:tswg1} line 5 and $S_0$ be the one that maximizes $\frac{K-k}{K}f(S_k)$ (Algorithm~\ref{alg:tswg1} line 6). We have
\begin{align*}
R(S^\star) &=  \frac{K-|S^\star|}{K}\mathbb {E}_{Y_a\sim D_a}\Big[\max(\max_{a \in S^\star}Y_a, \max_{j \in A\backslash S^\star}\mu_j)\Big] \\
&\leq  \frac{K-|S^\star|}{K} \mathbb {E}_{Y_a\sim D_a}\Big[(\max_{a \in S^\star}Y_a + \max_{j \in A}\mu_j)\Big] \\
&\leq \frac{K-|S^\star|}{K} \mathbb {E}_{Y_a\sim D_a}\Big[\max_{a \in S^\star}Y_a\Big] + \max_{j \in A}\mu_j \\
&= \frac{K-|S^\star|}{K} f(S^\star)+ \max_{j \in A}\mu_j \\
&\overset{a}\leq \frac{K-|S^\star|}{K} \frac{e}{e-1}f(S_{|S^\star|})+ \max_{j \in A}\mu_j \\
&\overset{b}\leq \frac{K-|S_0|}{K}\frac{e}{e-1} f(S_0)+ \max_{j \in A}\mu_j \\
&\overset{c}\leq \frac{e}{e-1}  R(S^{pr}) + R(S^{pr}) \\
&= \frac{2e-1}{e-1}  R(S^{pr})
\end{align*}
\noindent where (a) is due to the fact that $f(\cdot)$ is monotone and submodular, (b) follows from the definition of $S_0$, and (c) is due to the way $S^{pr}$ is defined. 
\end{IEEEproof}

Since for each $k \in \{1,...,K-1\}$, the algorithm evaluates $f(S_{k-1} \bigcup\{a\})$ for each $a \in A\backslash S_{k-1}$, it involves $O(NK)$ evaluation of $f(\cdot)$ in total, which determines its time complexity.   

\begin{remark}
For Bernoulli arms, Algorithm~\ref{alg:tswg4} is optimal ($\alpha=1$). We omit the proof but note that it can be established using a similar argument as the proof of Theorem~\ref{thm:Bernoulli} in the adaptive setting. 
\end{remark}

\begin{remark}
The above result assumes that $f(S)$ can be accurately estimated for any $S$. For general distributions, $f(S)$ needs to be estimated using samples generated by $\mathbb{D}$. Thus, the performance of the algorithm will be affected by the number of samples used.  
\end{remark}



\begin{algorithm}[!t]
  \caption{Offline non-adaptive probing} 
  \label{alg:tswg1}
  \begin{algorithmic}[1]
    \Require
    $A$: a set of $N$ arms; $K$: probing budget; $\mathbb{D}$: reward distributions for all the arms
    \Ensure
    $S^{\rm pr}$: the probing set
    \For {$k=0$ to $K-1$}
    \State $S_k \leftarrow \emptyset$
    \EndFor
    \For {$k=1$ to $K-1$}
       \State $a \leftarrow \mathop{\rm argmax}\limits_{a \in A \backslash S_{k-1}}{(f(S_{k-1} \bigcup\{a\})-f(S_{k-1}))}$
       \State $S_k \leftarrow S_{k-1} \cup \{a\}$
    \EndFor
    \State $S^{\rm pr} \leftarrow \mathop{\rm argmax}\limits_{k}{\frac{K-k}{K}f(S_k)}$
    \State $a \leftarrow \mathop{\rm argmax}\limits_{a \in A}\mu_{a}$
    \If{$f(S^{pr}) < \mu_a$}
        $S^{pr} \leftarrow \emptyset$
    \EndIf
  \end{algorithmic}
\end{algorithm}

\subsection{The offline problem with adaptive probing}


We then consider the more challenging setting with adaptive probing. 
The offline problem of finding the optimal $\pi_1$ can be formulated as a finite horizon Markov decision process (MDP) by viewing the set of probed arms and their outcomes (i.e., $\{a_1,...,a_i, Y_{a_1},...Y_{a_i}\}$ by the $i$-th probing step) as a state and the arm to be probed next or stopping probing as an action. The main challenge, however, is introduced by the combinatorial nature of the problem, leading to large state and action spaces. To derive a more scalable solution, we first observe that once the ordering of the set of arms to be probed is fixed, then an adaptive probing policy simplifies to a binary decision (whether to continue probing or not) at each probing step given the current observations. As we will see for Bernoulli arms (that is, $D_{a|x}$ is a Bernoulli distribution for all context-arm pairs), the natural ordering of probing the arms with larger mean values first is optimal. Although it is difficult to identify the optimal ordering for general reward distributions, the greedy ordering based on marginal improvement used in Algorithms~\ref{alg:tswg4} and~\ref{alg:tswg1} is a natural choice and achieves good performance as we show in the simulations. 

Below we assume that the set of arms $A$ has been sorted according to a certain ordering scheme and let $a_i$ denote the $i$-th arm on the list. 
We can then formulate the adaptive probing problem as an MDP with action space $=\{0,1\}$ with 1 indicating probing the next arm on the list and 0 indicating stopping probing. Further, the state at the $i$-th probing step can be defined as $(Y_1,...Y_i)$ as the set of arms probed is determined by $i$. To further simplify the problem, a key observation is that instead of keeping the complete outcomes, it suffices to maintain $s_i \triangleq \max \{Y_1,..., Y_i\}$ and define $(i,s_i)$ as the state of the $i$-th probing step. This can be proved by showing that the MDP using the compressed state space has the same optimal state-action values as the full MDP. We skip the formal proof due to the space constraint. 

With the above discussion, the adaptive probing problem can be reduced to a finite horizon MDP with a state space of size $K \times |\mathcal{Y}|$ (the number of possible probing outcomes for each arm) and an action space of size 2. Further, an adaptive probing policy $\pi_1$ is a mapping from each state $(k,s_k)$ to the action space. Let $V^*_i(s_i)$ denote the optimal state value at the $i$-th probing step under state $s_i$ (the maximum value observed so far). We have the following Bellman optimality equation:
\begin{align}
V^\star_i(s_{i}) &= \max\Big\{R_i(s_i),\sum\limits_{s_{i+1} \in \mathcal{Y}}{\rm{Pr}}(s_{i+1}|s_{i})V_{i+1}^\star(s_{i+1})\Big\}\label{bell}
\end{align}
where $R_i(s_i) \triangleq \frac{K-i}{K}\max(s_i, \max\limits_{j> i}\mu_j)$ is the value obtained if we stop probing after step $i$, and ${\rm{Pr}}(s_{i+1}|s_{i})$ is the probability that the next state becomes $s_{i+1}$ when the current state is $s_i$ and probing continues, which can be derived from $\mathbb{D}$. It is clear that $V^\star_K(s_{K}) = 0$ for any $s_K \in \mathcal{Y}$. The optimal value (under the fixed ordering) is then given by $V^\star_0(0)$ and the optimal probing action under each state can be derived by comparing the two terms in~\eqref{bell}.

Based on Equation (\ref{bell}), the optimal probing policy (under a fixed ordering of arms) can be computed using a dynamic programming based solution (see Algorithm \ref{alg_new}). 
The algorithm first sorts all the arms according to a certain ordering scheme. In this work, we assume that the arms are sorted greedily in terms of the marginal improvement with respect to $f(\cdot)$ as in the non-adaptive setting (i.e., lines 2-4 in Algorithm~\ref{alg:tswg4} with $k=K$). 
After initializing the values at the $K$-th probing step, the algorithm computes $V_k^\star(s_{k})$ and the policy $\pi_1((k,s_k)) \in \{0,1\}$ for $k = K-1,...,1$ recursively and for $s_k \in \mathcal{Y}$, using Equation~\eqref{bell}.  

\begin{algorithm}[!t]
  \caption{Offline adaptive probing via dynamic programming} 
  \label{alg_new}
  \begin{algorithmic}[1]
    \Require
    $A$: a set of $N$ arms; $K$: the probing budget; $\mathbb{D}$: distribution for all the arms. 
    \Ensure
    $v$: state value function, $\pi_1$: an adaptive probing policy 
    \State Sort the set of arms greedily in terms of marginal improvement with respect to $f(\cdot)$
    \State $v_K(s_K) \leftarrow 0$ for all $s_K \in \mathcal{Y}$. 
    \For{$k=K-1$ to $1$}
    \For{each outcome $s_k \in \mathcal{Y}$} 
    \State $R_k(s_k) \leftarrow \frac{K-k}{K}\max(s_k,  \max\limits_{j > k}\mu_j)$
    \State $v_k(s_k) \leftarrow \max\{R_k(s_k), \mathbb{E}_{s_{k+1} \in \mathcal{Y}}[v_{k+1}(s_{k+1})]\}$
    \State $\pi_1(k,s_k) \leftarrow 1$
    \If {$v_k(s_k)=R_k(s_k)$}
    \State $\pi_1(k,s_k) \leftarrow 0$
    \EndIf
    \EndFor
    \EndFor
    \State $v_0(0)=\max\{\max\limits_{a \in A}\mu_a,\mathbb{E}_{s_1 \in \mathcal{Y}}[v_{1}(s_1)]\}$
    \ignore{
    \For {$k=0$ to $K'$}
    \If{$ac_k=0$}
    \State $k' \leftarrow k$
    \State $S' \leftarrow Se[k']$, $Se[k']$ means first $k'$ sequence of $Se$
    \State Break
    \EndIf
    \EndFor
    }
  \end{algorithmic}
\end{algorithm}

Since for each $k \in \{1,...,K-1\}$ and each possible outcome in $\mathcal{Y}$, the computation of $V_k(s_k)$ involves taking the maximum over unprobed arms (line 5) and taking the expectation over all possible outcomes in the next step (line 6), the time complexity of the algorithm is $O(|\mathcal{Y}|K(N+|\mathcal{Y}|))$.

\subsection{Adaptive probing for Bernoulli arms}
Although it is difficult to derive the performance bound of Algorithm~\ref{alg_new} for general $\mathbb{D}$, we now show that the algorithm is optimal for Bernoulli arms where $D_a$ is a Bernoulli distribution for all $a \in A$. To this end, we first show that it is optimal to probe the set of arms according to their mean reward $\mu_a$ non-increasingly (note that for Bernoulli arms, this ordering is equivalent to the default ordering we use in Algorithm~\ref{alg_new}). To see this, we first note that for Bernoulli arms, whenever $Y_{a_i} = 1$ is observed, the decision maker can stop probing and play $a_i$. With this observation, any adaptive probing policy can be defined as follows. First pick a sequence $\vec{S}=(a_1,...,a_k)$ of $k$ arms; (2) probe arms following the sequence until either a value of one is observed or $k$ continuous zeros are observed. Let $a_{k+1}$ denote the arm played in the latter case (i.e., the one with maximum $\mu$ in the remaining arms). Thus, any adaptive policy can be fully characterized by an arbitrary sequence $\vec{S}$ of any length $k \in \{0, K-1\}$. 
Let $R(\vec{S})$ denote the expected value by following $\vec{S}$.


\begin{theorem}\label{thm:Bernoulli}
For Bernoulli arms, it is optimal to probe the set of arms in their mean reward $\mu$ non-increasingly. 
\end{theorem}
\begin{IEEEproof}
Consider a general adaptive policy defined by $\vec{S}$ with $k$ arms. We can derive its expected value as follows. 

\begin{align*}
R(\vec{S})\overset{a}=&\mu(a_{1})\frac{K-1}{K}+(1-\mu(a_{1}))\mu(a_{2})\frac{K-2}{K}\\
&+(1-\mu(a_{1}))(1-\mu(a_{2}))\mu(a_{3})\frac{K-3}{K}+...\\
&+(1-\mu(a_{1}))...(1-\mu(a_{k-1}))\mu(a_{k})\frac{K-k}{K}\\
&+(1-\mu(a_{1}))...(1-\mu(a_{k}))\mu(a_{k+1})\frac{K-k}{K}\\
\overset{b}=&\frac{K-k}{K}[1-\prod\limits_{i=1}^{k+1}(1-\mu(a_{i}))]\\
&+\frac{1}{K}[1-\prod\limits_{i=1}^{k-1}(1-\mu(a_{i})]\\
&+\frac{1}{K}[1-\prod\limits_{i=1}^{k-2}(1-\mu(a_{i}))]+...\\
&+\frac{1}{K}[1-(1-\mu(a_{1}))(1-\mu(a_{2}))]\\
&+\frac{1}{K}\mu(a_{1})
\end{align*}
\noindent where (a) follows from the definition of the policy and (b) is based on the following observation. Note that the first term in the second expression corresponds to the expected reward obtained when at least one of the $k+1$ arms (including the played one) gives a one, assuming that the unit reward is $(K-k)/K$. However, when an arm is sampled to one before step $k$, their rewards are higher, which are compensated by the following terms. The second expression makes it clear that is always better to choose arms with larger mean values for probing and play. To further show that it is optimal to probe arms in a non-increasingly order, assume that $\mu_{a_i} < \mu_{a_{i+1}}$ for some $i$ in the sequence. Let $\vec{S}'$ denote the sequence constructed from $\vec{S}$ by switching $a_i$ and $a_{i+1}$ while keeping the ordering of other arms. From the first expression above, we have  
\begin{align*}
R(\vec{S}')-R(\vec{S}) &= \prod^{i-1}_{j=1}(1-\mu_{a_j})\frac{K-i}{K}[\mu_{a_{i+1}}-\mu_{a_i}] \\ &\hspace{2ex}+\prod^{i-1}_{j=1}(1-\mu_{a_j})\frac{K-(i+1)}{K}\Big[(1-\mu_{a_{i+1}})\mu_{a_{i}}\\
&\hspace{25ex}-(1-\mu_{a_{i}})\mu_{a_{i+1}}\Big]\\
&= \prod^{i-1}_{j=1}(1-\mu_{a_j})\frac{K-i}{K}[\mu_{a_{i+1}}-\mu_{a_i}] \\ &\hspace{2ex}+\prod^{i-1}_{j=1}(1-\mu_{a_j})\frac{K-(i+1)}{K}\Big[\mu_{a_{i}}-\mu_{a_{i+1}}\Big]\\
&\geq 0
\end{align*}
\end{IEEEproof}

From Theorem~\ref{thm:Bernoulli}, it is optimal to probe arms with non-decreasing $\mu$. Thus, Algorithm~\ref{alg_new} is optimal for Bernoulli arms.

\section{The Online Setting}
In this section, we study the online setting where $\mathbb{D}$ is unknown a priori. We show that the offline algorithms developed in the previous section can be integrated with the combinatorial bandit framework in~\cite{chen2016combinatorial} to solve the joint probing and play problem in the online setting. We further establish the regret of our solution under general arm reward distributions when probing is non-adaptive and for Bernoulli arms when probing is adaptive.

\subsection{A combinatorial contextual bandit based algorithm}

We first present our algorithm for the online setting (see Algorithm \ref{alg:tswg2}). Our solution is adapted from the combinatorial bandit algorithm in~\cite{chen2016combinatorial}, which was originally developed for general reward functions without considering context. In contrast, we consider the contextual setting and further distinguish the different reward functions introduced by non-adaptive and adaptive probing.  

The algorithm (Algorithm \ref{alg:tswg2}) first normalizes the context space to $X'$ (see the discussion below). For each context-arm pair $a \in A$ and $x \in X'$, the algorithm maintains the number of times that arm $a$ has been probed or played under context $x$ as $n_{a|x}$, and an approximation $\tilde{F}_{a|x}$ of the true cumulative distribution function (CDF) $F_{a|x}$ of $D_{a|x}$. Note that $F_{a|x}$ with support $\{y_{{a|x},1},y_{{a|x},2},...,y_{{a|x},l}\}$ is defined as $F_{a|x}(y_{a,j})=\mathop{\rm Pr}\limits_{Y_{a|x} \sim D_{a|x}}[Y_{a|x} \le y_{{a|x},j}]$. 
Since $F_{a|x}$ is unknown, we approximate it by $\tilde{F}_{a|x}(y)$, which is computed as the fraction of the observed outcomes from arm $a$ under context $x$ that are no larger than $y$.

Algorithm \ref{alg:tswg2} plays each arm once under each context, and updates $n_{a|x}$ and $\tilde{F}_{a|x}$. After that, it maintains a lower bound 
$\underline{F}_{a|x}$ of $\tilde{F}_{a|x}$ using a UCB type of technique (lines 11-13). 
Let $\underline{D}=\underline{D}_{1|x} \times \underline{D}_{2|x} \times \cdots \times \underline{D}_{N|x}$ denote the corresponding distributions across all arms. The algorithm then distinguishes two cases. When non-adaptive probing is used, it invokes Algorithm~\ref{alg:tswg1} with $\underline{D}$ as the input. Otherwise, it invokes Algorithm~\ref{alg_new}. In each case, a probing set $S^{\rm pr}$ can be derived by following the policy obtained. We then collect samples for arms in $S^{\rm pr}$. For arms not in $S^{\rm pr}$, we estimate their mean reward using $\underline{D}_{a|x}$. We then identify the arm to play according to the play strategy $\pi_2$ given in Observation~\ref{obs:pi2}, and update $n_{a|x}$ and $\tilde{F}_{a|x}$ based on the probing and play results (lines 23-28). It is crucial to note that the play decision is based on $\underline{D}$ for unprobed arms (line 24) rather than the true distributions $\mathbb{D}$, which introduces another level of suboptimality. 

\begin{algorithm}[!t]
  \caption{Online contextual bandit for joint probing and play} 
  \label{alg:tswg2}
  \begin{algorithmic}[1]
    \Require
    $A$: a set of $N$ arms; $X$: a set of context; $K$: probing budget; $T$: time horizon
    \State $X' \leftarrow$ the normalization of context space  $X$. 
    \State $n_{a|x}$ $\leftarrow$ $0$, $\tilde{F}_{a|x}$ $\leftarrow$ $0$ for all $a \in A, x \in X'$
    \ignore{
    \For {each context $x \in X'$}
    \State $n_{a|x}$ $\leftarrow$ $0$, $\tilde{F}_{a|x}$ $\leftarrow$ $0$
    \For {$a=1$ to $N$}
    \State Play arm $a$, get $Y_{a|x}$
    \State $n_{a|x}=n_{a|x}+1$
    \State 
    \parbox[t]{\dimexpr\linewidth-\algorithmicindent}
    {Update $\tilde{F}_{a|x}$ based on $\{y_{{a|x},1},y_{{a|x},2},...,y_{{a|x},l}\}$ and $Y_{a|x}$}
    \EndFor
    \EndFor
    }
    \For {$t=1,2,...,T$}
       \State Receive context $x_t \in X$
       \State $x \leftarrow \bar{x}_t$
       \If{there is an arm $a$ never played under $x$}
    \State Play arm $a$, get $Y_{a|x}$
    \State $n_{a|x}=n_{a|x}+1$
    \State 
    \parbox[t]{\dimexpr\linewidth-\algorithmicindent}
    {Update $\tilde{F}_{a|x}$ based on $Y_{a|x}$}
    \State \textbf{continue} 
       \EndIf
       \State \parbox[t]{\dimexpr\linewidth-\algorithmicindent}{
       For each arm $a \in A$, let $\underline{D}_{a|x}$ be a distribution whose CDF $\underline{F}_{a|x}$ is given by}
       \State \parbox[t]{\dimexpr\linewidth-\algorithmicindent}{\begin{align*}
\underline{F}_{a|x}=
\left\{
             \begin{array}{ll}
              {\rm max}\{\tilde{F}_{a|x}(y)-\sqrt{\frac{3{\rm ln}t}{2n_{a|x}}},0\}, &0 \le y < 1  \\
              \\
              1,   &y=1 
             \end{array}
\right.
\end{align*} }
       \State $\underline{D}=\underline{D}_{1|x} \times \underline{D}_{2|x} \times \cdots \times \underline{D}_{N|x}$
       \If {probing is non-adaptive}
       \State $S_t^{\rm pr} \leftarrow {\rm Algorithm \ref{alg:tswg1}}(A, K, \underline{D})$
       \State Probe each arm $a \in S_t^{\rm pr}$ and get $Y_{a|x}$
    \State \Call 
      {Play and Update}{$S_t^{\rm pr}, \{Y_{a|x}: a \in S_t^{\rm pr}\}, x$}
      
    \EndIf
    \If {probing is adaptive}
    \State $\pi_1 \leftarrow {\rm Algorithm \ref{alg_new}}(A, K, \underline{D})$
    \State \parbox[t]{\dimexpr\linewidth-\algorithmicindent}{Probe arms according to $\pi_1$ and get their outcomes}
    \State $S_t^{\rm pr} \leftarrow $ the set of arms probed
    \State \Call 
      {Play and Update}{$S_t^{\rm pr}, \{Y_{a|x}: a \in S_t^{\rm pr}\},x$}
    \EndIf
    \EndFor
    \Function 
    {Play and Update}{$S_t^{\rm pr}, Y_t, x$}:
       \State \parbox[t]{\dimexpr\linewidth-\algorithmicindent}
    {
   Play $a_t^{\rm pl} \leftarrow \mathop{\rm argmax}\limits_a \{\{Y_{a|x}: a \in S_t^{\rm pr}\}, \{\mathbb{E}(\underline{D}_{a|x}): a \in A \backslash S_t^{\rm pr}\}\}$}
   \If {$a_t^{\rm pl} \notin S_t^{\rm pr}$} get $Y_{a_t^{\rm pl}|x}$
   \EndIf
    \For {each $j \in \{S_t^{\rm pr}, a_t^{\rm pl}\}$} \State $n_{j|x}=n_{j|x}+1$ 
    \State \parbox[t]{\dimexpr\linewidth-\algorithmicindent}{Update $\tilde{F}_{j|x}$ based on  $\{y_{{j|x},1},y_{{j|x},2},...,y_{{j|x},l}\}$ \\ and $Y_{j|x}$}
    \EndFor
    \EndFunction
    
  \end{algorithmic}
\end{algorithm}

\subsection{Dealing with large context space}
When the context space is large or continuous, we group similar context together. Given $X =[0,1]$, let $X'$ denote the $\epsilon$-uniform mesh of $X$. When a new context $x_t \in X$ is received, 
it is mapped to the closet point in $X'$ according to
\begin{align*}
\bar{x}_t=\mathop{\rm argmin}\limits_{x \in X'}|x-x_t|
\end{align*}
where we choose the smaller value when there is a tie. 

To understand the loss introduced by the compression of the context space, let $(S^\star(x),a^\star(x))$ denote the optimal offline solution under context $x$ and $r(S^\star(x),a^\star(x)|x)$ its expected value when the solution is applied under context $x$. We define the discretization error as
\ignore{
We define the discretized best response:
\begin{align*}
S_X^\star(x)=S^\star(f_X(x)), \forall x \in X
\end{align*}
Then regret of ${\rm ALG}_X$ and
\red{
\begin{align*}
R_X(T)=\sum_{t=1}^T (r(S_X^\star(x_t)| f_X(x_t)) - r'(S_t, a^{\rm pl}_t| f_X(x_t)))
\end{align*}
}
}
\begin{align*}
DE(T)=\sum_{t=1}^T [r(S^\star(x_t),a^\star(x_t)|x) - r(S^\star(\bar{x}_t), a^\star(\bar{x}_t)|\bar{x})]
\end{align*}
We have the following lemma:
\begin{lemma}
\begin{align*}
DE(T) \le \epsilon LT
\end{align*}
\label{lemma_dis}
\end{lemma}
\begin{IEEEproof}
For each round $t$ and the respective context $x_t$, we have
\begin{align*}
r(S^\star(\bar{x}_t),a^\star(\bar{x}_t)|\bar{x}_t) &\ge r(S^\star(x_t),a^\star(x_t)|\bar{x}_t)\\
&\ge  r(S^\star(x_t),a^\star(x_t)| x_t) - \epsilon L 
\end{align*}
where the first inequality is because that $(S^\star(\bar{x}_t),a^\star(\bar{x}_t))$ is the optimal solution under context $\bar{x}_t$ and the second inequality is due to the definition of $\bar{x}_t$ and the Lipschtz-continuity assumption~\eqref{Lipschitzcontext}. Summing this up over all rounds $t$, we obtain
\begin{align*}
DE(T) \le \epsilon LT
\end{align*}
\end{IEEEproof}

With the above lemma, we can bound the total regret as 
\begin{align*}
\textsf{Reg}_{\alpha}(T) \leq \alpha\epsilon LT + R_{\alpha}(T)
\end{align*}
\noindent where $R_{\alpha}(T)$ is defined as
\begin{align*}
R_{\alpha}(T)=&\sum_{t=1}^T (\alpha r(S_t^\star,a^\star| \bar{x}_t) - r(S_t, a_t|  \bar{x}_t))
\end{align*}

\subsection{Theoretical analysis of the online algorithm}
In this section, we analyze the regret of Algorithm~\ref{alg:tswg2} for the non-adaptive probing setting under a general $\mathbb{D}$ and as well as the adaptive probing setting for Bernoulli arms. To simplify the notation, we let $R(S,Y,a|x)$ denote the value obtained given the probing and play decision $(S,a)$, the probing result $Y$, and context $x$, and $r(S,a|x)$ its expectation over $Y$ in both settings. It is understood that in the adaptive setting, $S$ is interpreted as a sequence with a pre-defined ordering (which is without loss of generality for Bernoulli bandits according to Theorem~\ref{thm:Bernoulli}). Note that although in the offline setting, the optimal $a$ is completely determined by $S$ and $Y$, the online algorithm picks $a$ according to $\underline{D}_{a|x}$. We include $a$ in the value functions to emphasize this fact. In both cases, we can show that $R(S,Y,a|x)$ satisfies the following properties required by ~\cite{chen2016combinatorial}. 


\begin{itemize}
    \item There exists $M>0$ such that for any $S$, $Y$, and $x$, we have $0 \le R(S,Y,a|x) \le M$. 
    \item $R(S,Y|x) \le R(S, Y'|x)$ for any $S$, $x$, $Y$ and $Y'$ such that $Y_i \leq Y'_i$ for every $a_i \in S$. 
\end{itemize}

The following definitions are adapted from~\cite{chen2016combinatorial}. For any $S \subseteq A$ and $a \in A$, we define 
\begin{align*}
\Delta_{S, a|x}=\max\{\alpha r(S^\star,a^\star|x) - r(S, a|x), 0\}
\end{align*}

Let $E_B(x)$ denote the set of arms that are contained in at least one suboptimal probing decision $S$ under context $x$. For each $i \in E_B(x)$, we define
\begin{align*}
\Delta(i,{\rm min}|x)={\rm min}\{ \Delta_{S, a|x}|S \subseteq A, a\in A, i \in S\}
\end{align*}
\begin{align*}
L_e={\rm max}\{ |E_B(\bar{x})||\bar{x} \in X'\}
\end{align*}
\noindent and 
\begin{align*}
\Delta({\rm min})={\rm min}\{\Delta(i,{\rm min}|\bar{x})| i \in E_B(\bar{x}), \bar{x}\in X' \}
\end{align*}

With the above definitions, we can prove the following regret bounds for both non-adaptive probing under general distributions and adaptive probing under Bernoulli distributions. 

\ignore{
\begin{assumption}
(Independent outcomes from arms). The outcomes from all $N$ arms are mutually
independent.
\end{assumption}
\begin{assumption}
There exists $M>0$ such that for any $y \in [0, 1]^N$ and any $S$ under any context $x$, we have $0 \le R(y, S|x) \le M$.
\end{assumption}
\begin{assumption}
If two vectors $y, y' \in [0, 1]^N$ satisfy $y_i \le y_i'(\forall i \in N)$, and for any $S$, under some context $x$, we have $R(y, S|x) \le R(y', S|x)$.
\end{assumption}
\red{
We define 
\begin{align}
\Delta_{S, a^{\rm pl}|x}=\mathop{\rm max}\{\alpha r(S^\star|x) - r'(S,, a^{\rm pl}|\bar{x}), 0\}
\end{align}
}
Let $E_B$ be the set of arms that are contained in at least one bad super arm. For each $i \in E_B$, we define
\red{
\begin{align}
\Delta(i,{\rm min}|\bar{x}_t)={\rm min}\{ \Delta_{S, a^{\rm pl}|x}|S \subseteq A, a^{\rm pl} \in A, i \in S \}
\end{align}
}
}

\begin{figure*}[t]
 	\vspace{-10pt}
 	\centering
 		\subfloat[]{%
 		\includegraphics[width=0.21\textwidth]{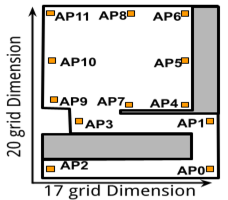}%
 		\label{student center}
 	}
 	\subfloat[]{%
 		\includegraphics[width=0.25\textwidth]{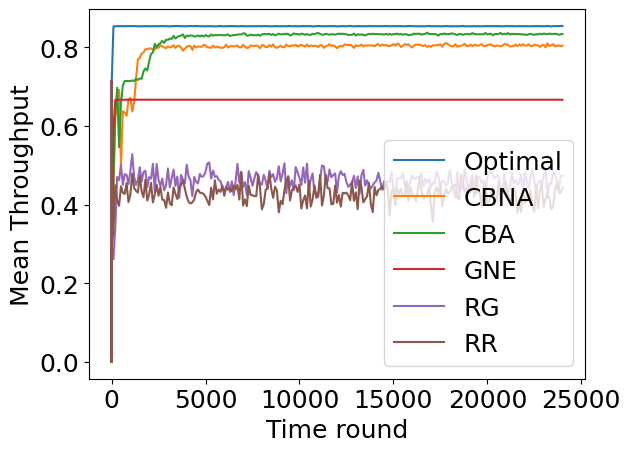}%
 		\label{fig1}
 	}
 	\subfloat[]{%
 		\includegraphics[width=0.25\textwidth]{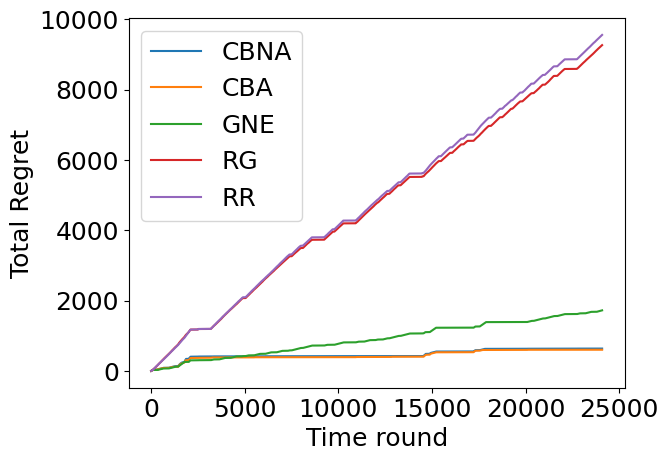}%
 		\label{fig2}
 	}
 	\subfloat[]{%
 		\includegraphics[width=0.25\textwidth]{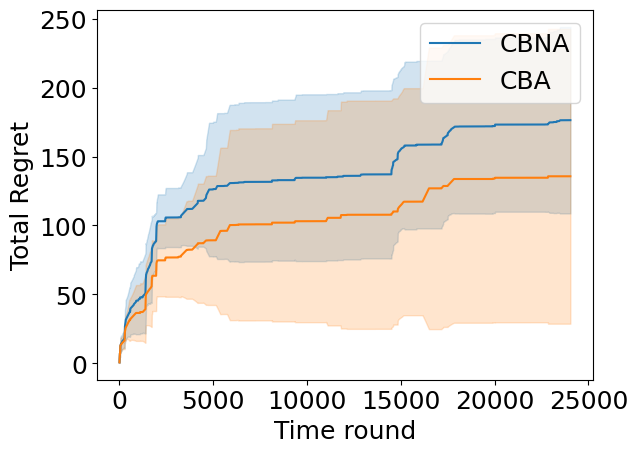}%
 		\label{fig3}
 	}
        \vspace{-5pt}

 	 \caption{\small (a) measurement setup (in a student hall); (b) mean throughput for Bernoulli arms: $K = 10, N = 10$, single context; (c) accumulative regret for Bernoulli arms (w.r.t. the optimal offline adaptive algorithm): $K = 10, N = 10, \frac{1}{\epsilon}=8$; (d) average accumulative regret for arms with general reward distributions (w.r.t. the optimal offline adaptive algorithm): $K = 4, N = 4, \frac{1}{\epsilon}=8$.
   }
 	\vspace{-15pt}
 	\label{fig}
 \end{figure*}

\begin{theorem}\label{thm:regret}
Algorithm~\ref{alg:tswg2} has the following distribution-dependent upper bound on the $\alpha$-approximation regret in $T$ rounds
\begin{align}
2[M^2(K-1)L_e\frac{2136}{\Delta({\rm min})}{\rm ln}T + (\frac{\pi^2}{3}+1)\alpha MN]^\frac{1}{2}(\alpha LT)^\frac{1}{2}
\label{regbound}
\end{align}

\noindent and the following distribution-independent upper bound on the $\alpha$-approximation regret in $T$ rounds 
\begin{align*}
2[93M\sqrt{N(K-1)T{\rm ln}T}+(\frac{\pi^2}{3}+1)\alpha MN]^\frac{1}{2}(\alpha LT)^\frac{1}{2}
\end{align*}

For Bernoulli arms, the above bounds can be simplified by taking $\alpha=1$ since the offline algorithms are optimal for both non-adaptive and adaptive settings for Bernoulli arms.

\label{theofinal}
\end{theorem}


To adapt the proof of Theorem 1 in \cite{chen2016combinatorial} to our problem, we note that Lemmas 1-3 in \cite{chen2016combinatorial} still hold in our setting (in the presence of context and joint probing and play) given the two properties of $R(S,Y,a|x)$ described above. Below we extend Lemma 4 in \cite{chen2016combinatorial}, which is the key lemma for proving the main result. 
\begin{lemma}
Let $\mathcal{T}$ denote the set of rounds $t$ over $T$ where every arm has been played at least once before round $t$ under the same context $\bar{x}_t$. 
Define an event in each round $t \in \mathcal{T}$: 
\begin{align*}
H_t=\Big\{0<\Delta_{S_t, a_t|\bar{x}_t}<4M\sum\limits_{a_t \in S_t}\sqrt{\frac{3{\rm ln}t}{2n_{a_t|\bar{x}_t}}}\Big\}
\end{align*}
Then 
\begin{align*}
R_{\alpha}(T) \leq \mathbb{E}[\sum_{\bar{x} \in X'}\sum_{t \in \mathcal{T},\bar{x}_t = \bar{x}} 
\mathbf{1}_{H_t} \Delta_{S_t, a_t|\bar{x}_t}]+(\frac{\pi^2}{3}+1)\alpha MN\frac{1}{\epsilon}
\end{align*}
where $\mathbf{1}$ is the indicator function. 
\label{5}
\end{lemma}
\begin{IEEEproof}
Let $F_{a|\bar{x}}$ be the CDF of $D_{a|\bar{x}}$ and $\tilde{F}_{a|\bar{x}, n_{a|\bar{x}}}$ the corresponding empirical CDF of the first $n_{a|\bar{x}}$ observations from arm $a$ under context $\bar{x}$. For each $t\in \mathcal{T}$, define an event
\begin{align*}
\mathfrak{C}_t&=\{\exists a \in A \ {\rm such \ that} \\ 
&\hspace{2ex}\mathop{\rm sup}\limits_{y \in [0, 1]}|\tilde{F}_{a|\bar{x}_t,n_{a|\bar{x}_t}}(y)-F_{a|\bar{x}_t}(y)| \ge \sqrt{\frac{3{\rm ln}t}{2n_{a|\bar{x}_t}}}\}
\end{align*}
We bound the $\alpha$-approximation regret of Algorithm~\ref{alg:tswg2} as
\begin{align}
R_{\alpha}(T)=&\sum_{t=1}^T (\alpha r(S_t^\star,a^\star| \bar{x}_t) - r(S_t, a_t| \bar{x}_t)) \nonumber \\ \le& \sum_{\bar{x} \in X'}
\sum_{t=1,\bar{x}_t = \bar{x}}^T\mathbb{E}[\Delta_{S_t, a_t|\bar{x}_t}] \nonumber \\
\label{17}
=&\mathbb{E}[\sum_{\bar{x} \in X'}
\sum_{t \not\in \mathcal{T},\bar{x}_t = \bar{x}}^T
\Delta_{S_t, a_t|\bar{x}_t}] \nonumber\\
&+\mathbb{E}[\sum_{\bar{x} \in X'}
\sum_{t \in \mathcal{T},\bar{x}_t = \bar{x}}\mathbf{1}\{\mathfrak{C}_t\}\Delta_{S_t, a_t|\bar{x}_t}] \nonumber \\
&+\mathbb{E}[\sum_{\bar{x} \in X'}
\sum_{t \in \mathcal{T},\bar{x}_t =\bar{x}}\mathbf{1}\{\mathfrak{\neg C}_t\}\Delta_{S_t, a_t|\bar{x}_t}]
\end{align}
The first term in (\ref{17}) can be bounded as
\begin{align*}
\mathbb{E}[\sum_{\bar{x} \in X'}
\sum_{t \not\in \mathcal{T},\bar{x}_t = \bar{x}}^T
\Delta_{S_t, a_t|\bar{x}_t}] &\le \sum_{\bar{x} \in X'}
\sum_{t \not\in \mathcal{T},\bar{x}_t = \bar{x}}^T\\
&\hspace{2ex}\alpha r(S_t^\star,a_t^\star| \bar{x}_t) \nonumber \\ &\leq \alpha MN\frac{1}{\epsilon}
\end{align*}
For the second term, by the DKW inequality of Lemma 2 in \cite{chen2016combinatorial}, we first have
\begin{align*}
{\rm Pr}[|\tilde{F}_{a|x_q,q}-F_{a|x_q}| \ge \sqrt{\frac{3{\rm ln}t}{2q}}] \le 2e^{-2q\frac{3{\rm ln}t}{2q}} = 2t^{-3}
\end{align*}
Now we have
\begin{align*}
\mathbb{E}[\sum_{\bar{x} \in X'}
\sum_{t \in \mathcal{T},\bar{x}_t = \bar{x}}\mathbf{1}\{\mathfrak{C}_t\}] \le& \sum_{\bar{x} \in X'}
\sum_{t \in \mathcal{T},\bar{x}_t = \bar{x}}^T\sum_{a \in A}\sum_{q < t, \bar{x}_q = \bar{x}}^t\\
&{\rm Pr}[|\tilde{F}_{a|\bar{x}_q,q}-F_{a|\bar{x}_q}| \ge \sqrt{\frac{3{\rm ln}t}{2q}}]\\
\le&\sum_{\bar{x} \in X'}
\sum_{t \in \mathcal{T},\bar{x}_t = \bar{x}}^T\sum_{a \in A}\sum_{q < t, \bar{x}_q = \bar{x}}^t2t^{-3}\\
\le& 2N\frac{1}{\epsilon}\sum_{t \in \mathcal{T},\bar{x}_t = \bar{x}}^T t^{-2}\\
\le& \frac{\pi^2}{3}N\frac{1}{\epsilon}
\end{align*}
So the second term can be bounded as
\begin{align*}
\mathbb{E}[\sum_{\bar{x} \in X'}
\sum_{t \in \mathcal{T},\bar{x}_t = \bar{x}}\mathbf{1}\{\mathfrak{C}_t\}\Delta_{S, a|\bar{x}_t}] &\le \frac{\pi^2}{3}N\frac{1}{\epsilon}\alpha r(S_t^\star| \bar{x}_t) \\ &\le \frac{\pi^2}{3}N\frac{1}{\epsilon}\alpha M
\end{align*}
For the third term, we can directly apply the bound for a single context in Lemma 4 of \cite{chen2016combinatorial}.
So finally we have 
\begin{align*}
R_{\alpha}(T) \le \mathbb{E}[\sum_{\bar{x} \in X'}\sum_{t \in \mathcal{T},\bar{x}_t = \bar{x}} 
\mathbf{1}_{H_t} \Delta_{S_t, a_t|\bar{x}_t}]+(\frac{\pi^2}{3}+1)\alpha MN\frac{1}{\epsilon}
\end{align*}
\end{IEEEproof}

The rest of the proof follows the proof of Theorem 1 in \cite{chen2016combinatorial}.
Then we can get
\begin{align*}
R_{\alpha}(T) \leq& M^2(K-1)\frac{1}{\epsilon}L_e\frac{2136}{\Delta({\rm min})}{\rm ln}T \\
&+ (\frac{\pi^2}{3}+1)\alpha MN\frac{1}{\epsilon}
\end{align*}
By applying Lemma~\ref{lemma_dis}, we have
\begin{align*}
\textsf{Reg}_{\alpha}(T) \leq& M^2(K-1)\frac{1}{\epsilon}L_e\frac{2136}{\Delta({\rm min})}{\rm ln}T \\
&+ (\frac{\pi^2}{3}+1)\alpha MN\frac{1}{\epsilon}+\alpha\epsilon LT.
\label{regbound2}
\end{align*}


\noindent By letting $\epsilon = \frac{[M^2(K-1)L_e\frac{2136}{\Delta({\rm min})}{\rm ln}T + (\frac{\pi^2}{3}+1)\alpha MN]^\frac{1}{2}}{(\alpha LT)^\frac{1}{2}}$, we have the following distribution-dependent upper bound on the $\alpha$-approximation regret in $T$ rounds
\begin{align*}
\textsf{Reg}_{\alpha}(T) \leq& 2[M^2(K-1)L_e\frac{2136}{\Delta({\rm min})}{\rm ln}T \\
&+ (\frac{\pi^2}{3}+1)\alpha MN]^\frac{1}{2}(\alpha LT)^\frac{1}{2}.
\end{align*}

Similarly, we have the following distribution-independent upper bound on the $\alpha$-approximation regret in $T$ rounds
\begin{align*}
\textsf{Reg}_{\alpha}(T) \le& 93M\sqrt{\frac{1}{\epsilon}N(K-1)T{\rm ln}T}\\
&+(\frac{\pi^2}{3}+1)\alpha MN\frac{1}{\epsilon}+\alpha\epsilon LT \\
\le& 93M\frac{1}{\epsilon}\sqrt{N(K-1)T{\rm ln}T}\\
&+(\frac{\pi^2}{3}+1)\alpha MN\frac{1}{\epsilon}+\alpha\epsilon LT.
\end{align*}


\noindent Setting $\epsilon = \frac{[93M\sqrt{N(K-1)T{\rm ln}T}+(\frac{\pi^2}{3}+1)\alpha MN]^\frac{1}{2}}{(\alpha LT)^\frac{1}{2}}$, we have
\begin{align*}
\textsf{Reg}_{\alpha}(T) \le& 2[93M\sqrt{N(K-1)T{\rm ln}T}\\
&+(\frac{\pi^2}{3}+1)\alpha MN]^\frac{1}{2}(\alpha LT)^\frac{1}{2}.
\end{align*}

\section{Evaluation}
In this section, we evaluate the performance of our algorithms using simulations with channel and mobility traces collected from a real testbed. 

\subsection{Baselines}
We consider the following algorithms in the evaluation. For non-adaptive probing, we apply Algorithm~\ref{alg:tswg2} together with  Algorithm~\ref{alg:tswg1} and call it the contextual bandit with non-adaptive probing algorithm (CBNA). For adaptive-probing, we apply Algorithm~\ref{alg:tswg2} together with Algorithm~\ref{alg_new}, 
which is referred as the contextual bandit with adaptive probing algorithm (CBA). We use 50 samples to evaluate $f(\cdot)$ in Algorithm~\ref{alg:tswg1}.

We compare our algorithms with four baselines: (a) \textsf{Optimal}, which is the optimal offline solution under adaptive probing and is obtained through an exhaustive search over all possible ordering of arms in Algorithm~\ref{alg_new}. (b) GNE (greedy probing and greedy play without exploration in the play), which follows CBNA except that at the play stage, $\underline{F}_{a|x_t}$ is replaced with $\tilde{F}_{a|x_t}$ so that no exploration is considered in play decisions. (c) RG (randomly probing and greedy play) where the decision maker randomly selects a $k$ and probes $k$ arms randomly selected. The play policy is the same as CBNA. (d) RR (random probing and random play) where the decision maker randomly selects a $k$ and probes $k$ arms randomly selected. It then plays a randomly selected arm.

\subsection{Evaluation Settings}
We collect channel traces from a real testbed using an 802.11ad router and laptop, as well as a commercially available mmWave channel simulator (Remcom Insite \cite{remcom}) to evaluate the system. We collect SNR traces with actual testbeds in 250 different locations in the student hall (Scenario Fig.~\ref{student center}) with a particle size of 0.8m. In the student hall, 12 Airfide \cite{airfide} 802.11ad APs are installed, each with a 64-sector 8-phased array antenna. Clients are Acer TravelMate-648 \cite{acer-P648} laptops with a single phased array and 36 sectors. 
We modified the open-source driver\cite{wil6210} on both APs and the client to extract SNR and beamforming information. At each grid, the client conducted beamforming with all APs and recorded the per-sector SNR. We repeated this measurement ten times at each grid and stored the average SNR for different Tx and Rx sectors in a $64 \times 36$ matrix.

We use the 802.11ad \cite{80211ad} MCS-SNR table to map the SNR and get the average throughput of the connection (as in \cite{mobihoc18-mmchoir}) after obtaining the signal strength channel trace from the testbed and Remcom channel simulator. It is based on the best Tx/Rx sector pair of links discovered through the beamforming process. In each context $x$ (client location), we normalize the average throughput of each AP $a$ as $\mu(a|x)$ and consider two settings for link rate distributions. In the Bernoulli setting, the probability of getting a reward of 1 from arm $a$ under context $x$ is $\mu(a|x)$, and the probability of getting zero reward is $1-\mu(a|x)$. We also consider a more general setting where the link rates are supported on $\mathcal{Y}=[0, \frac{1}{3}, \frac{2}{3}, 1]$. For each AP $a$ and location $x$, we randomly assign the probabilities to the four points so that the mean link rate is equal to $\mu(a|x)$.

For mobility traces, we choose 15-80 customers randomly located and configure their walking patterns by observing the typical walking behavior of a room. We assume that all clients have a walking speed of $1 m/s $. At any time, a client is in one of the grids where the channel is measured by the testbed or channel simulator as described above. We choose 10 clients' traces in the simulation, where each trace has 80 steps and each step includes 30 time rounds. We then implemented all the algorithms of AP selection with python based on the collected traces and showed the results below.

\subsection{Results}

We evaluate all the algorithms with different probing budget $K$ and number of APs $N$. We randomly select $N$ APs and compare the algorithms in three different settings. 

In Fig.~\ref{fig1}, we consider the Bernoulli setting under a single context, where we randomly pick a location and keep it the same all the time (without mobility). We plot the mean throughput averaged over 100 time rounds. We observe that although GNE and CBNA obtain best performance in the first $\sim2000$ time rounds, CBA surpasses them after that and obtains higher average throughput. Additionally, we observe that CBNA and CBA consistently outperform the others. Also, the performance of GNE is most stable due to the lack of exploration.

In Fig.~\ref{fig2}, we consider the Bernoulli setting with multiple contexts. We evenly divide the room into a grid of 8 cells and treat each of them as a context. 
For each location $x$ visited by a client, we map $x$ to the center of its closest cell to get $\bar{x}$. Fig.~\ref{fig2} plots the expected accumulative expected regret of each baseline with respect to \textsf{Optimal}. We find that the total expected regret of CBNA and CBA is less than other baselines. We further notice that in this figure, there are two jumps around time rounds 15,000 and 17,000, respectively, which correspond to the two starting points that have not been explored enough before, such as the bottom part where most APs are blocked. In addition, we find that for Bernoulli arms, the regret of CBNA and CBA is relatively close, and CBA is slightly better than CBNA. 

In order to further distinguish the two algorithms, we plot the results for the general distribution setting described above in Fig.~\ref{fig3}. Here we again group locations into 8 context cells as in Fig.~\ref{fig2}. 
We run both algorithms five times with different random seeds. The area between the upper blue curve and lower gray curve indicates the standard deviation of CBNA, and the area between the upper grey curve and the lower yellow curve indicates the standard deviation of CBA. We observe that CBA always performs better than CBNA in the simulated scenarios, indicating the importance of considering adaptive probing.


\section{Conclusion}
In this paper, we propose an online learning based joint beamforming and scheduling framework for throughput optimization for mmWave WLANs. We model the problem as a contextual bandit problem with joint probing and play. In the offline setting with known link rate distributions, we develop an approximation algorithm when the probing decision is non-adaptive and a dynamic programming based algorithm for the adaptive setting. Both algorithms are optimal for Bernoulli link rates and we establish an approximation factor of the former under general link rate distributions. We further develop a combinatorial contextual bandit algorithm for the online setting and prove its regret. Our algorithms are validated through simulations with channel and mobile traces collected from a real testbed.   

\section*{Acknowledgment}
This work was supported in part by NSF grant CNS-1816943.

\bibliographystyle{ieeetr} 
\bibliography{references}
\end{document}